\pdfoutput=1
\documentclass[11pt]{article}
\usepackage[]{acl}

\usepackage{times}
\usepackage{latexsym}

\usepackage[T1]{fontenc}
\usepackage[utf8]{inputenc}

\usepackage{microtype}

\usepackage{color}
\usepackage{amsmath}
\usepackage{graphicx}
\usepackage{arydshln} % dashline
\usepackage{multirow}

\definecolor{c1}{HTML}{95bddc}
\definecolor{c2}{HTML}{c2d1e5}
\definecolor{c3}{HTML}{fe793d}
\definecolor{c4}{HTML}{fb4c1f}
\definecolor{c5}{HTML}{b71a3b}
\definecolor{c6}{HTML}{7e0f12}
\definecolor{c7}{HTML}{E85642}
\definecolor{c8}{HTML}{C00000}
\definecolor{c9}{HTML}{ff2d51}
\definecolor{blue}{HTML}{339ff4}
\definecolor{green}{HTML}{3ca057}
\title{Exploring Information Processing in Large Language Models: \\Insights from Information Bottleneck Theory}

\author{
Zhou Yang$^{\dagger}$, Zhengyu Qi$^{\ddagger}$, Zhaochun Ren$^{\ddagger}$, Zhikai Jia$^{\mathsection}$\\
\textbf{Haizhou Sun$^{\dagger\dagger}$}, \textbf{Xiaofei Zhu$^{\mathparagraph}$},
\textbf{Xiangwen Liao}$^\dagger$\thanks{\hspace{1mm} Corresponding author.}\\
\small $^\dagger$College of Computer and Data Science, Fuzhou University,
 Fuzhou, China; 
\small $^\ddagger$Leiden University, Leiden, The Netherlands; \\
\small $^\mathsection$ SCITIX (SGP) TECH PTE. LTD, Singapore;
\small $^{\dagger\dagger}$H. Sun is with SmartMore, Shenzhen, China; \\
\small $^\mathparagraph$College of Computer Science and Technology, Chongqing University of Technology, Chongqing, China \\ 
\small  \texttt{200310007@fzu.edu.cn, liaoxw@fzu.edu.cn} \\
}

\begin{document}
\maketitle

\begin{abstract}
Large Language Models (LLMs) have demonstrated remarkable performance across a wide range of tasks by understanding input information and predicting corresponding outputs. 
However, the internal mechanisms by which LLMs comprehend input and make effective predictions remain poorly understood. 
In this paper, we explore the working mechanism of LLMs in information processing from the perspective of Information Bottleneck Theory. 
We propose a non-training construction strategy to define a task space and identify the following key findings: (1) LLMs compress input information into specific task spaces (e.g., sentiment space, topic space) to facilitate task understanding; (2) they then extract and utilize relevant information from the task space at critical moments to generate accurate predictions. Based on these insights, we introduce two novel approaches: an Information Compression-based Context Learning (IC-ICL) and a Task-Space-guided Fine-Tuning (TS-FT). IC-ICL enhances reasoning performance and inference efficiency by compressing retrieved example information into the task space. TS-FT employs a space-guided loss to fine-tune LLMs, encouraging the learning of more effective compression and selection mechanisms. Experiments across multiple datasets validate the effectiveness of task space construction. Additionally, IC-ICL not only improves performance but also accelerates inference speed by over 40\%, while TS-FT achieves superior results with a minimal strategy adjustment \footnote{The entire development process relies on the Siflow platform (https://scitix.ai/), provided by SCITIX (SGP) TECH PTE. LTD.}
\footnote{Our code will be released soon.}.
\end{abstract}

\section{Introduction}
Large Language Models (LLMs) have achieved remarkable success in natural language processing (NLP), demonstrating exceptional performance across a wide range of tasks such as text generation, machine translation, and sentiment analysis. By understanding input information and predicting corresponding outputs, LLMs have shown strong capabilities in handling complex tasks. However, despite their impressive real-world performance, the internal mechanisms by which LLMs comprehend input and make accurate predictions remain largely unexplored.

In this paper, we investigate the information processing mechanisms of Large Language Models (LLMs) from the perspective of Information Bottleneck Theory\footnote{Information Bottleneck Theory is a framework that optimizes the trade-off between retaining relevant information for a task and discarding redundant data by compressing the input.}. We propose a non-gradient-based task space detection strategy, which helps trace the internal information flow within LLMs. 
Using this strategy, we investigate the information flow across layers of LLMs during comprehension and prediction phase. 
During the understanding phase, LLMs compress input information into specific task spaces. 
In the prediction phase, LLMs extract and integrate relevant information from the task spaces, decompressing it at critical moments to generate predictions.
That is, LLMs perform task comprehension and prediction by compressing and decompressing information within specific task spaces.
Further results show that while LLMs effectively compress high-quality information during the compression phase, they struggle to decompress it during the prediction phase, leading to suboptimal performance.

Based on these insights, we propose two novel methods derived from Information Bottleneck Theory: Information Compression-based Context Learning (IC-ICL) and Task-Space-guided Fine-Tuning (TS-FT). IC-ICL retrieves relevant examples and maps them into the task space, enhancing LLMs' decompression capabilities to improve prediction accuracy. TS-FT constructs high-quality decompressed representations within the task space and employs a spatially-guided loss function to help LLMs learn better decompression strategies.

We validate our proposed methods through experiments on multiple datasets. The results demonstrate that IC-ICL significantly enhances reasoning accuracy while accelerating inference speed by over 40\%. TS-FT enhances model performance through simple fine-tuning without the need for complex adjustments.

Overall, our contributions are as follows:

\begin{itemize}
\item 
We introduce a non-gradient-based task space detection method that aids in detecting information flow changes.

\item 
We show that LLMs process tasks by compressing and decompressing information within task-specific spaces, exhibiting strong compression but weaker decompression abilities.

\item 
We introduce an Information Compression-based Context Learning method that substantially improves performance while accelerating inference speed by 40\%.

\item 
We present a Task-Space-guided Fine-Tuning method that enhances LLMs' information processing capabilities through a simple and effective space-guided loss function.

\end{itemize}

\section{Information Detection for LLMs}
In this section, we explore the information detection mechanisms within Large Language Models (LLMs) from the perspective of Information Bottleneck Theory, focusing on how information is processed and optimized during task comprehension and prediction.

\subsection{Task-Space-based Information Detection Strategy}
We propose a task-space-based strategy for detecting information flow in LLMs, which leverages the concept of compressing input information into task-specific spaces for efficient processing.

\begin{table*}
  \centering
  \caption{Prompts for the task space of the emotion ``joyful''}
  \begin{tabular}{p{5cm}p{9.5cm}}
      \hline
          Prompt Type      & Prompt Content \\
      \hline
          \multirow{3}{*}{\centering \shortstack{Positive-related Prompt}}      &  Infer the dialogue from the perspective of the emotion ``\textbf{joyful}''.\\ 
          & Dialogue Context: [Dialogue Context] \\ 
          & Response Format: ``Emotion: [Inferred Emotion]''  \\
          \hline
          \multirow{3}{*}{\centering \shortstack{Negative-related Prompt}}     &  Infer the dialogue from the perspective of the emotion ``\textbf{angry}''.\\ 
          & Dialogue Context: [Dialogue Context] \\ 
          & Response Format: ``Emotion: [Inferred Emotion]''  \\
      \hline
  \end{tabular}
  \label{table prompt 6.1}
\end{table*}

The task space is composed of multiple basic vectors, each constructed from the features that best represent the task at hand. For instance, in emotion classification, the most representative feature is the emotion categories. Therefore, we construct the emotional categories involved in the task as basic dimension vectors. Similarly, for topic classification task, we construct the topic types as the basic vectors.

For simplicity, we use emotion categories as an example to describe the construction process. For a emotion category $e_i \in E$, such as ``joyful'', we construct the input pairs shown in Table 1, i.e., positive-related prompt $P^+$ and negative-related prompt $P^-$. The negative-related prompt pairs consist of randomly sampled other emotion types, while the dialogue context is drawn from Empathetic Dialogues, containing dialogue statements with the emotion category $e_i$.

Based on these prompts, the LLM generates tokens $y^+_t$ and $y^-_t$ at time step $t$. 
\begin{gather}
    Y^+ = LLM(y^+_t|P^+, y^+_{<t}) \\
    Y^- = LLM(y^-_t|P^-, y^-_{<t})
\end{gather}

\begin{figure}
\centering
\includegraphics[width=70mm]{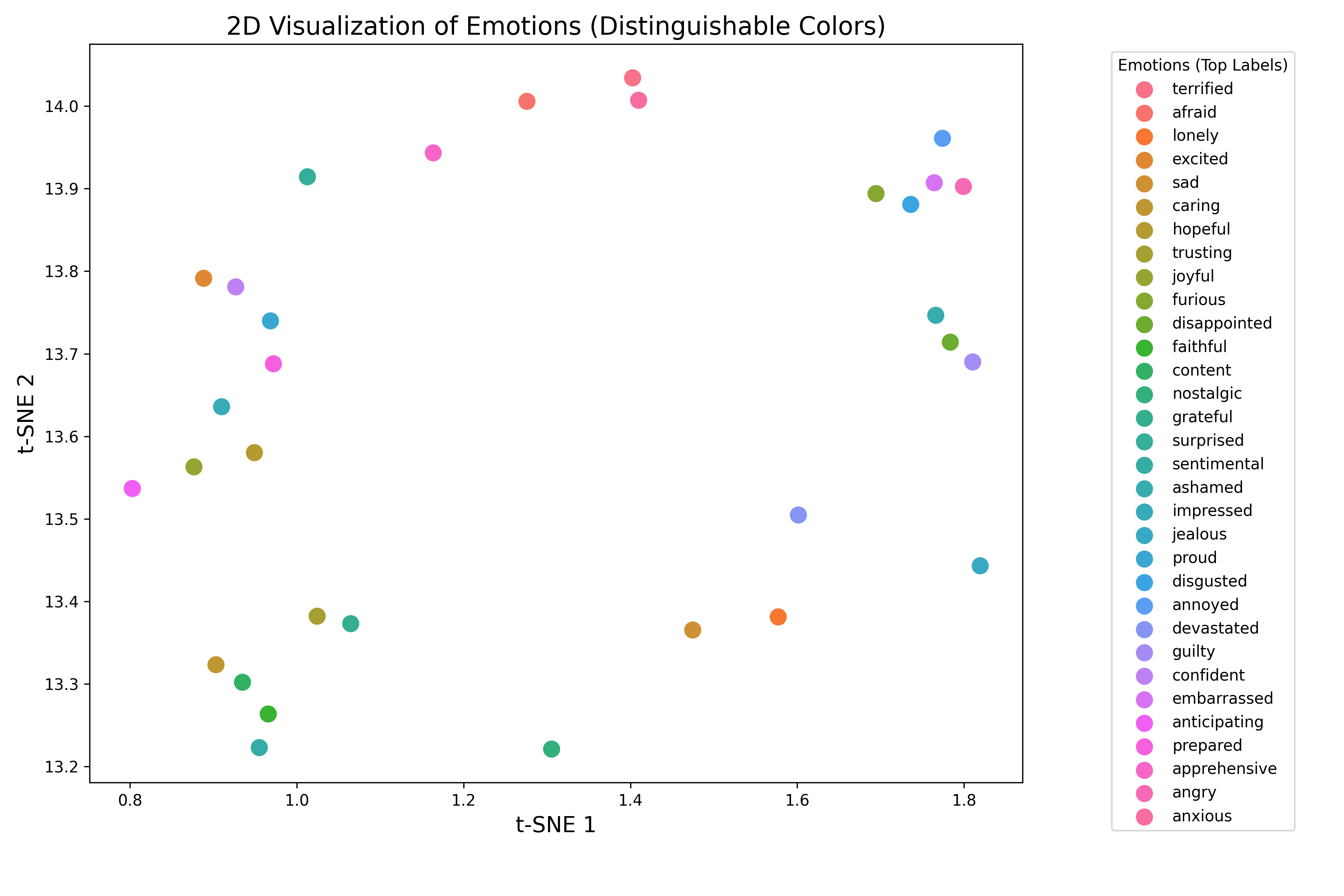}
\caption{\label{fig emotions_2d}
2D visualization of the emotion space.
}
\end{figure}

We then extract the hidden vectors at layer $l$ of the LLM, obtaining the positive-related hidden layer $h^+_{t,l}$ and the negative-related hidden vector representation $h^-_{t,l}$ for tokens $y^+_t$ and $y^-_t$ at time step $t$. Following prior methods~\cite{Liu2024CtrlAAR}, we obtain the directional hidden vector representation $H^l_t$ for time step $t$ at layer $l$ by subtracting the two vectors.
\begin{gather}
    h^l_t = h^+_{t, l} - h^-_{t, l}
\end{gather}

Using the above method, we collect $N_h$ direction vectors for the emotion type $e_i$. These vectors contain both the representation of the emotion type $e_i$ and noise. To purify the emotion type representation, Principal Component Analysis (PCA) is applied, resulting in refined direction vectors $H^l_{e_i}$ for $e_i$. In this paper, we treat these direction vectors as dimensional vectors of the space elements.
\begin{gather}
    H^l_{e_i} = PCA(H^l_{E_i})
    %\widetilde{}
\end{gather}

\subsection{Constructing Task Spaces}
The construction of task spaces is central to our approach, as it defines how information is organized and compressed to facilitate task comprehension and prediction.
To validate the rationality of the task space, we visualize it for observation and analysis.

\begin{figure}
\centering
\includegraphics[width=70mm]{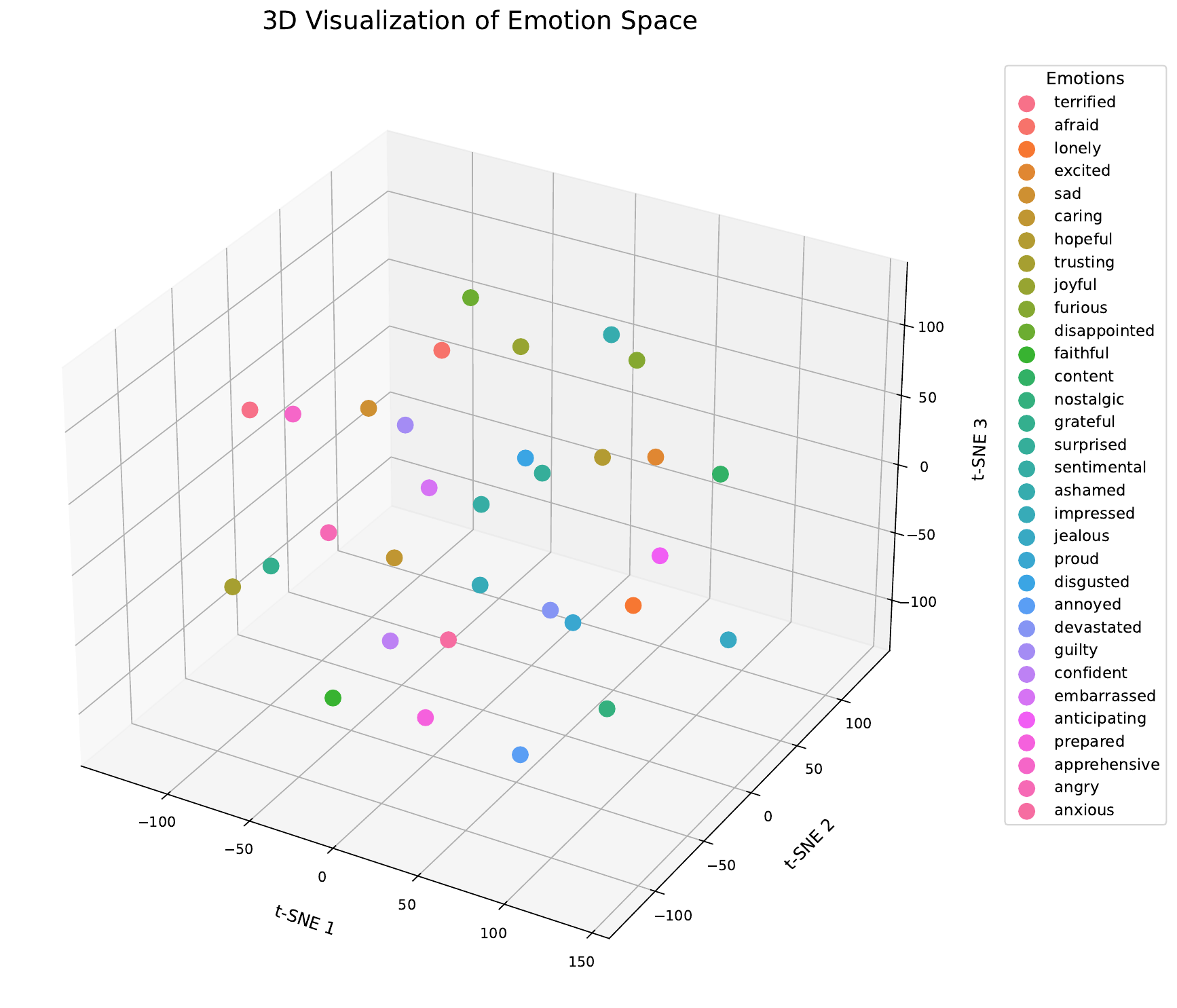}
\caption{\label{fig emotions_3d}
3D visualization of the emotion space.
}
\end{figure}

Figures \ref{fig emotions_2d} and \ref{fig emotions_3d} present the 2D and 3D visualizations of the emotion space after dimensionality reduction using t-SNE. The distribution of emotions in the figures shows that the emotions are fairly evenly distributed in the space, with similar emotions clustering closer together. For example, "terrified," "afraid," "anxious," and "angry" are located closer to each other.

\begin{figure}
\centering
\includegraphics[width=70mm]{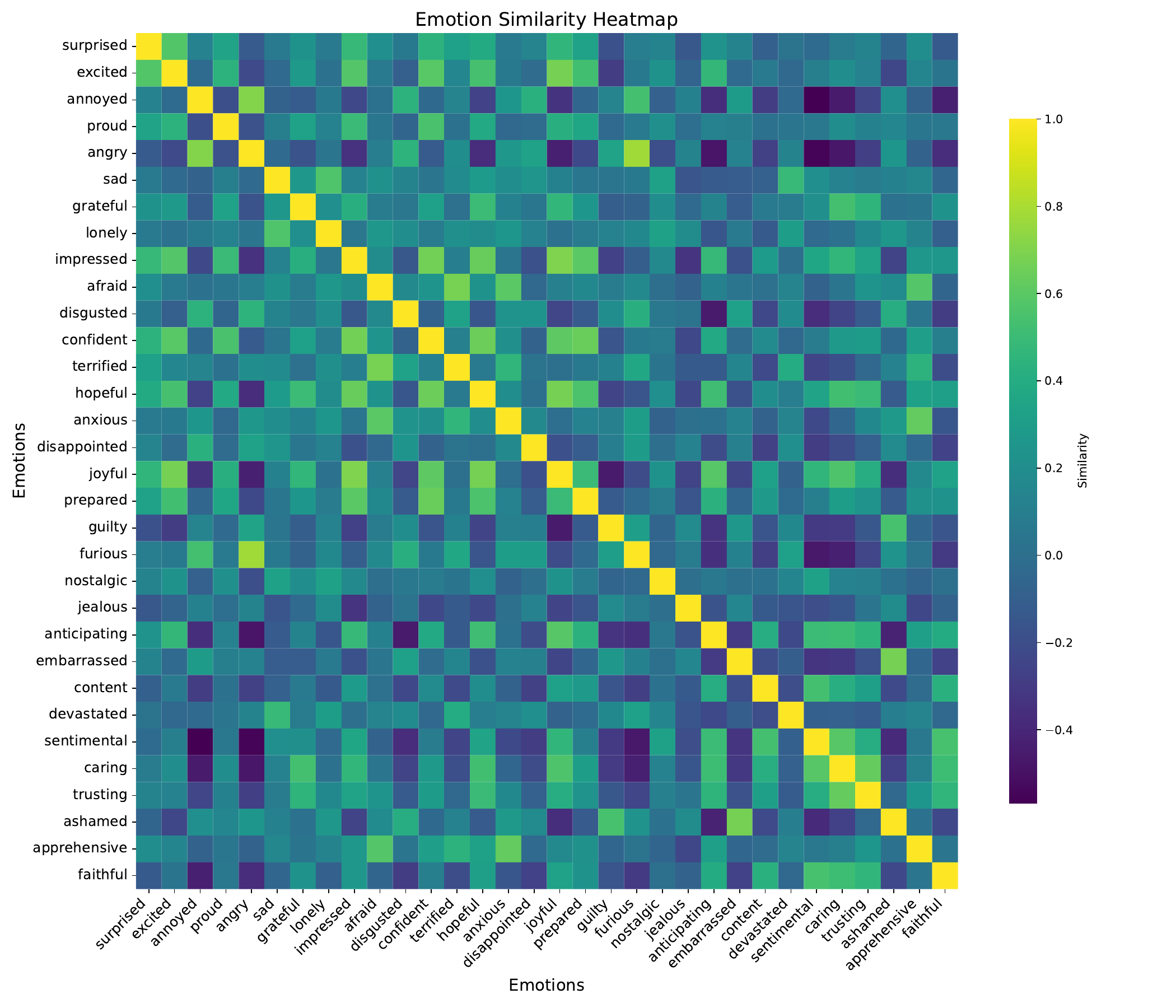}
\caption{\label{fig emotion_similarity_heatmap}
Visualization of emotion similarity
}
\end{figure}

To more clearly illustrate the similarity between emotions, we also construct a heatmap of the cosine distances between emotions. The heatmap, shown in Figure \ref{fig emotion_similarity_heatmap}, calculates the cosine distance between pairs of emotions, where a brighter color indicates a smaller distance (i.e., higher similarity). The results explicitly show that more similar emotions, such as ``excited'' and ``joyful'', have higher similarity scores, while more distinct emotions, such as ``annoyed'' and ``caring'', exhibit lower similarity.

Based on these analyses and observations, the distribution of emotion categories in the space appears to be both uniform and reasonable, supporting the validity of the emotion space.

\subsection{Information Compression in Understanding}
During the understanding phase, LLMs compress input information into task-specific spaces, retaining only the most relevant details for effective task processing.

\subsubsection{Calculation of Mutual Information}
\textbf{Objective}:
The goal of this experiment is to verify the information state of input samples in the emotion space.

\textbf{Hypothesis}:
The representation of the samples compresses toward a specific emotion space.

\textbf{Steps}:
(1) \textbf{Sample Representation}.
We selected $n$ samples $S$ from the Empathetic Dialogue dataset for mutual information statistical experiments.
For sample $s_i$, its hidden layer representation $h^l_s$ at the $l$-th layer is obtained by inputting it into the LLMs.

\begin{gather}
    h^l_{s_i}=[h^l_{0}, ...,h^l_{j}, ...,h^l_{N}] \\
    h^l_{s_i} \in R^{d \times N}
\end{gather}

(2) \textbf{Emotion Space Representation}.
For sample $s_i$, we define its ground-truth emotion label as $e_{s_i}$.
To observe the sample's state in the emotion space, we define multiple emotion spaces: $es^{top}{1}$, $es^{top}{2}$, $es^{top}{d_k}$, and $es^{top}{d_e}$.

\begin{gather}
    e_{top} = Top^{cosine}_{d_k}(e_{s_i}, e_j) \\
e_{d_k}= \frac{1}{d_k} \sum_{top_j=1}^{d_k} e_{top_j}
\end{gather}
Here, $\text{Top}^{cosine}_{d_k}$ is a function that sorts emotions based on cosine similarity and selects the top $d_k$ most similar emotion types. $e_j \in E$ represents the emotions from the set of all emotion types $E$.
When $d_k = 1$, the emotion space at the $l$-th layer corresponds to the base vector derived from the ground-truth emotion label. When $d_k \neq 1$, the emotion space at the $l$-th layer is the mean of the $d_k$ nearest neighbors (including the sample's own label) based on cosine distance.

(3) \textbf{Projection to Emotion Space}.
For the $j$-th token's representation at the $l$-th layer, $h^l_{j}$, we project it onto the corresponding emotion space $es^{l, top}_{d_k}$:
\begin{gather}
    h^{p,l}_j = \frac{h^l_{j} \cdot es^{l, top}_{d_k}}{|es^{l, top}_{d_k} \cdot es^{l, top}_{d_k}|} es^{l, top}_{d_k}
\end{gather}

For the $N$ tokens in sample $s_i$ during the comprehension process, we sum the projected representations $h^{p,l}_j$ to obtain the overall projection of the sample:
\begin{gather}
    h^{p,l}_{s_i} = \sum_{j=1}^{N} h^{p,l}_j
\end{gather}

\textbf{Mutual Information Estimation}.
Mutual Information (MI) measures the dependency between two random variables \(X\) and \(Y\). It quantifies the reduction in uncertainty of one variable given the other. The mutual information is defined as:
\begin{gather}
I(X; Y) = \int \int p(x, y) \log \frac{p(x, y)}{p(x)p(y)} \, dx \, dy,
\end{gather}
where \(p(x, y)\) is the joint probability density function of \(X\) and \(Y\), and \(p(x)\), \(p(y)\) are their respective marginal probability densities.

In practice, exact computation of mutual information is infeasible as the true probability distributions are unknown. Therefore, we employ K-Nearest Neighbors (KNN)-based methods to approximate the densities. Using these approximations, the MI can be expressed as:
\begin{gather}
I(X; Y) \approx \frac{1}{N} \sum_{i=1}^N \log \frac{\hat{p}(x_i, y_i)}{\hat{p}(x_i)\hat{p}(y_i)},
\end{gather}
where \(\hat{p}(x_i, y_i)\), \(\hat{p}(x_i)\), and \(\hat{p}(y_i)\) are the estimated joint and marginal probabilities for the samples \(x_i\) and \(y_i\). 

Using the method described above, the mutual information $I(h^{p,l}_{s_i}; es^{l, top}_{d_k})$  between the sample $s_i$ and the emotion space $es^{l, top}_{d_k}$ at the $l$-th layer can be obtained.
At the same time, we also computed the mutual information $I(h^{p,l}_{s_i}; h^{p,0}_{s_i})$ between the $l$-th layer and the $0$-th layer.

\subsubsection{Analysis of Results}

According to information bottleneck theory, information during the comprehension process should move away from the initial space and towards the target emotion space.
That is, the mutual information between the $l$-th layer's hidden state and the 0-th layer's hidden state should gradually decrease, while the mutual information with the target emotion space should gradually increase.

\begin{figure}
\centering
\includegraphics[width=70mm]{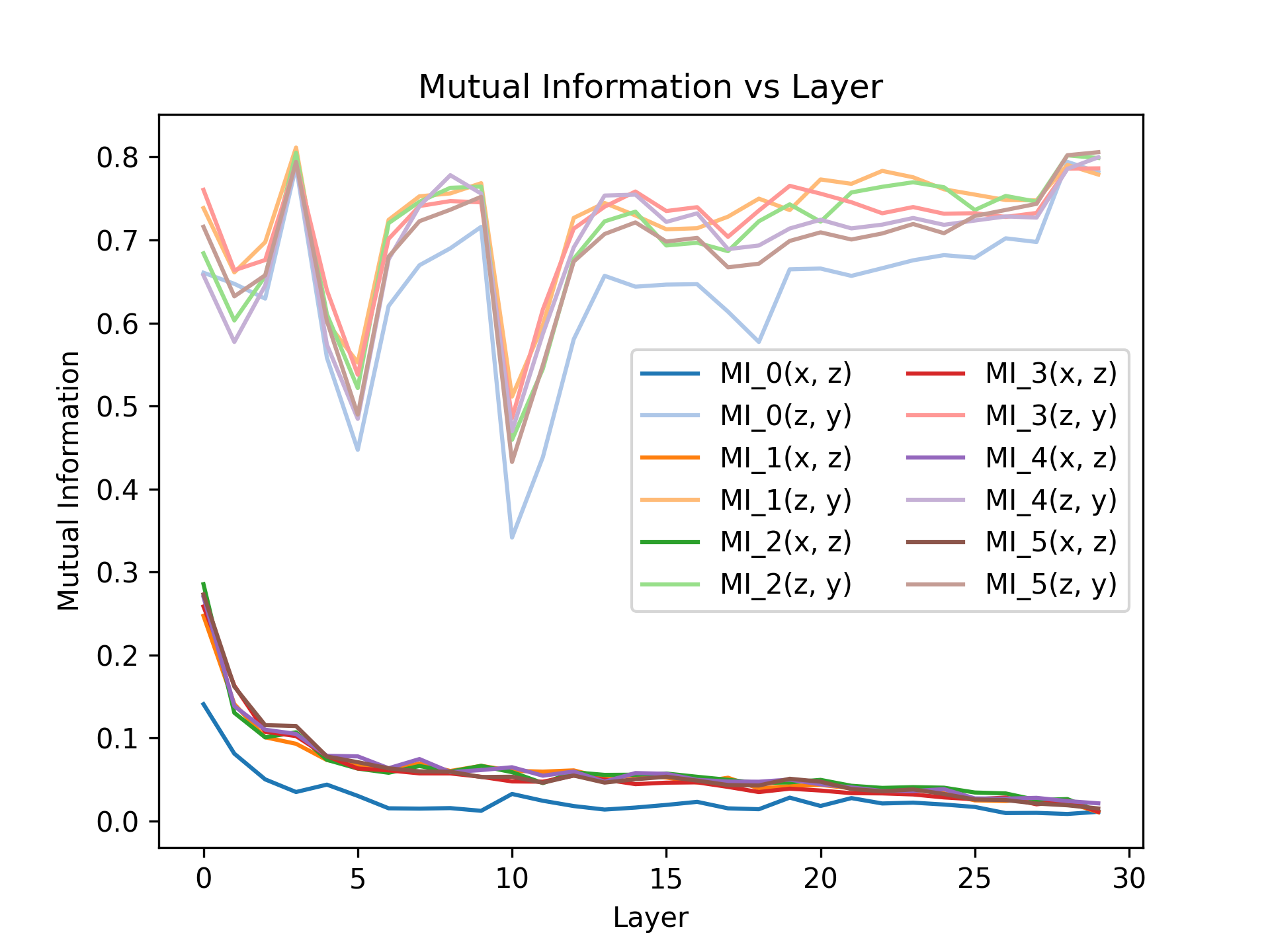}
\caption{\label{fig mi_top1}
Information variation of LLMs in the ground-truth emotion space.
}
\end{figure}

Figure \ref{fig mi_top1} shows the variation in mutual information of the 
$l$-th layer’s hidden state in the ground-truth emotion space, i.e., with 
$d_k=1$.
According to the results, the mutual information in the earlier layers of the LLM fluctuates significantly. This is primarily because the shallow layers of LLMs mainly process basic information such as syntax and grammar.
However, after layers 12 to 28, the mutual information between the hidden state $z$ and the input $x$ gradually decreases, while the mutual information with the target emotion space $y$ gradually increases.
This suggests that LLMs compress the input content towards a specific emotion space.

\begin{figure}
\centering
\includegraphics[width=70mm]{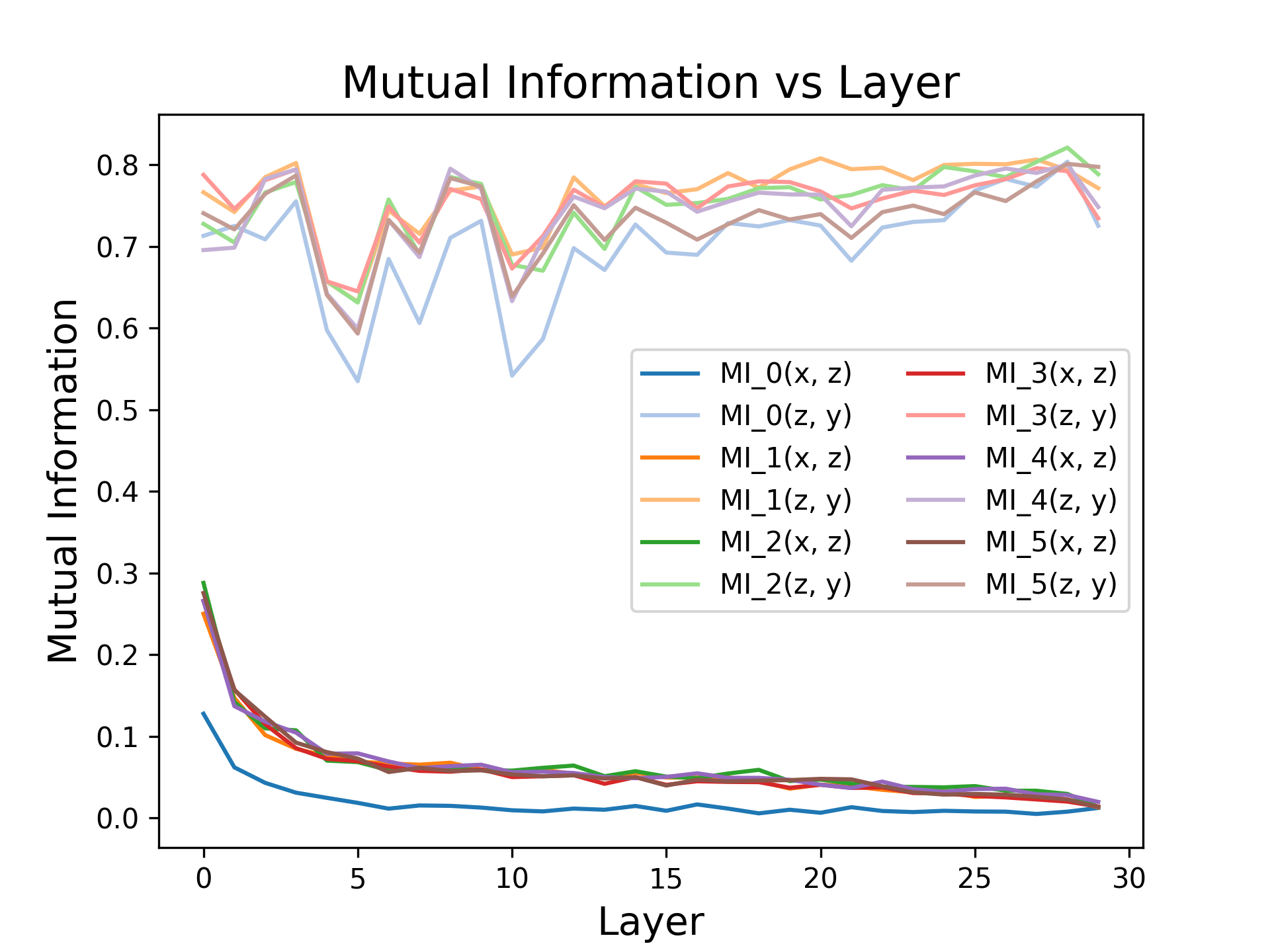}
\caption{\label{fig mi_top2}
Information Variation of LLMs in the Emotion Space with $d_k=2$
}
\end{figure}

\begin{figure}
\centering
\includegraphics[width=70mm]{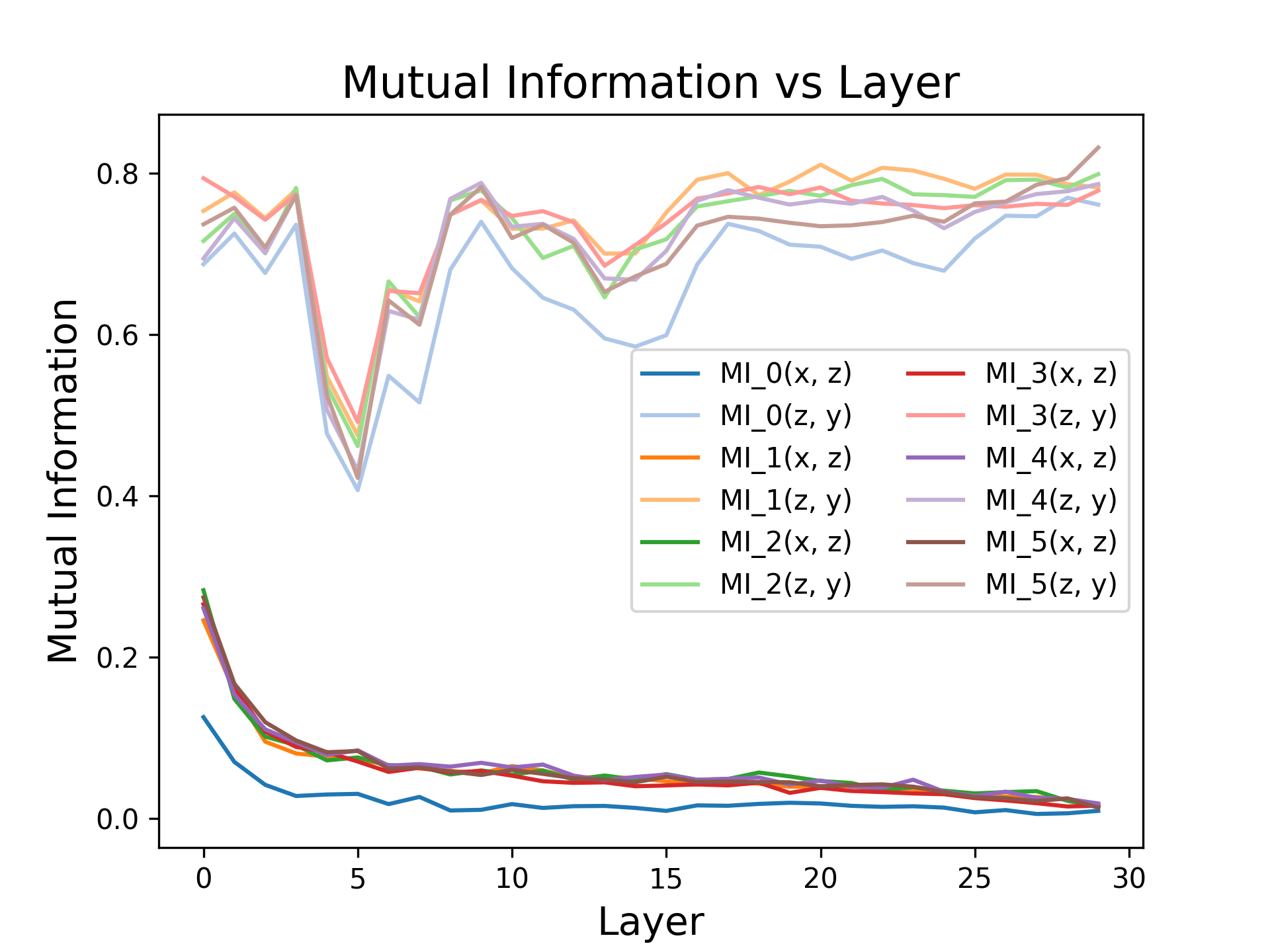}
\caption{\label{fig mi_top5}
Information Variation of LLMs in the Emotion Space with $d_k=5$
}
\end{figure}

Figures \ref{fig mi_top2} and \ref{fig mi_top5} show the variation in mutual information across a broader range of emotion spaces, i.e., with 
$d_k=1$.
The results also exhibit the same trend.

\begin{figure}
\centering
\includegraphics[width=70mm]{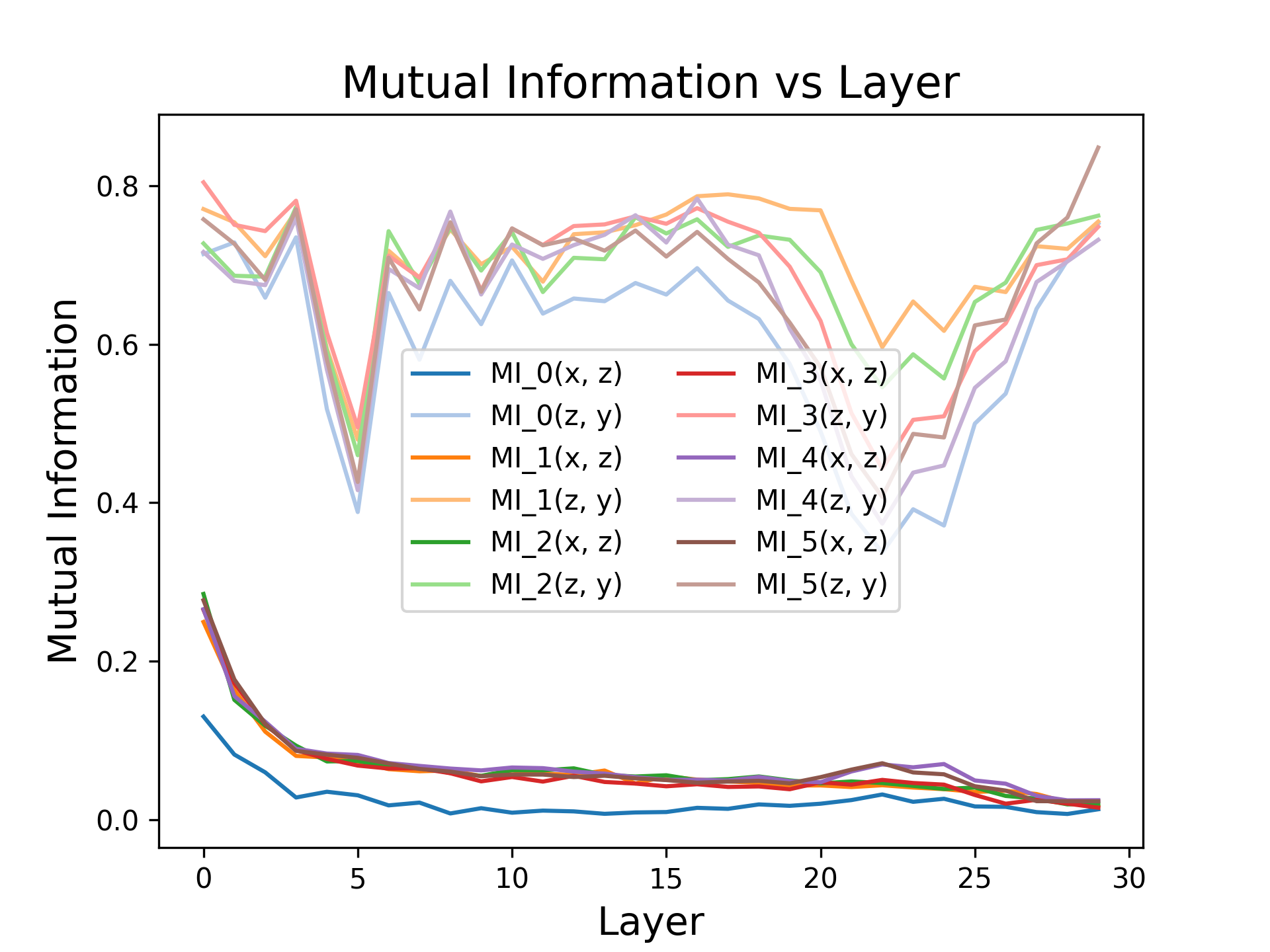}
\caption{\label{fig mi_top32}
Information Variation of LLMs in the Emotion Space with $d_k=32$
}
\end{figure}

Specifically, we plotted the mutual information variation for the widest emotion space, with 
$d_k=32$, as shown in Figure \ref{fig mi_top32}.
We observe that the mutual information continues to change in this space.
Combining data from all emotion spaces, this suggests that LLMs actually compress information towards an emotion space closer to the ground-truth, rather than compressing information across the entire emotion space.

% To estimate the densities, the kernel density estimation (KDE) method is utilized. The KDE for a single variable is defined as:
% \begin{gather}
% \hat{p}(x) = \frac{1}{Nh} \sum_{i=1}^N K\left(\frac{x - x_i}{h}\right),
% \end{gather}
% where \(K\) is the kernel function (e.g., Gaussian kernel), and \(h\) is the bandwidth parameter controlling the smoothness of the density estimation.

\subsection{Information Decompression in Prediction}
In the prediction phase, LLMs decompress the relevant information from task spaces to generate accurate predictions, highlighting the challenges and limitations in the decompression process.

\subsubsection{Information Measurement During the Prediction}

\begin{figure}
\centering
\includegraphics[width=70mm]{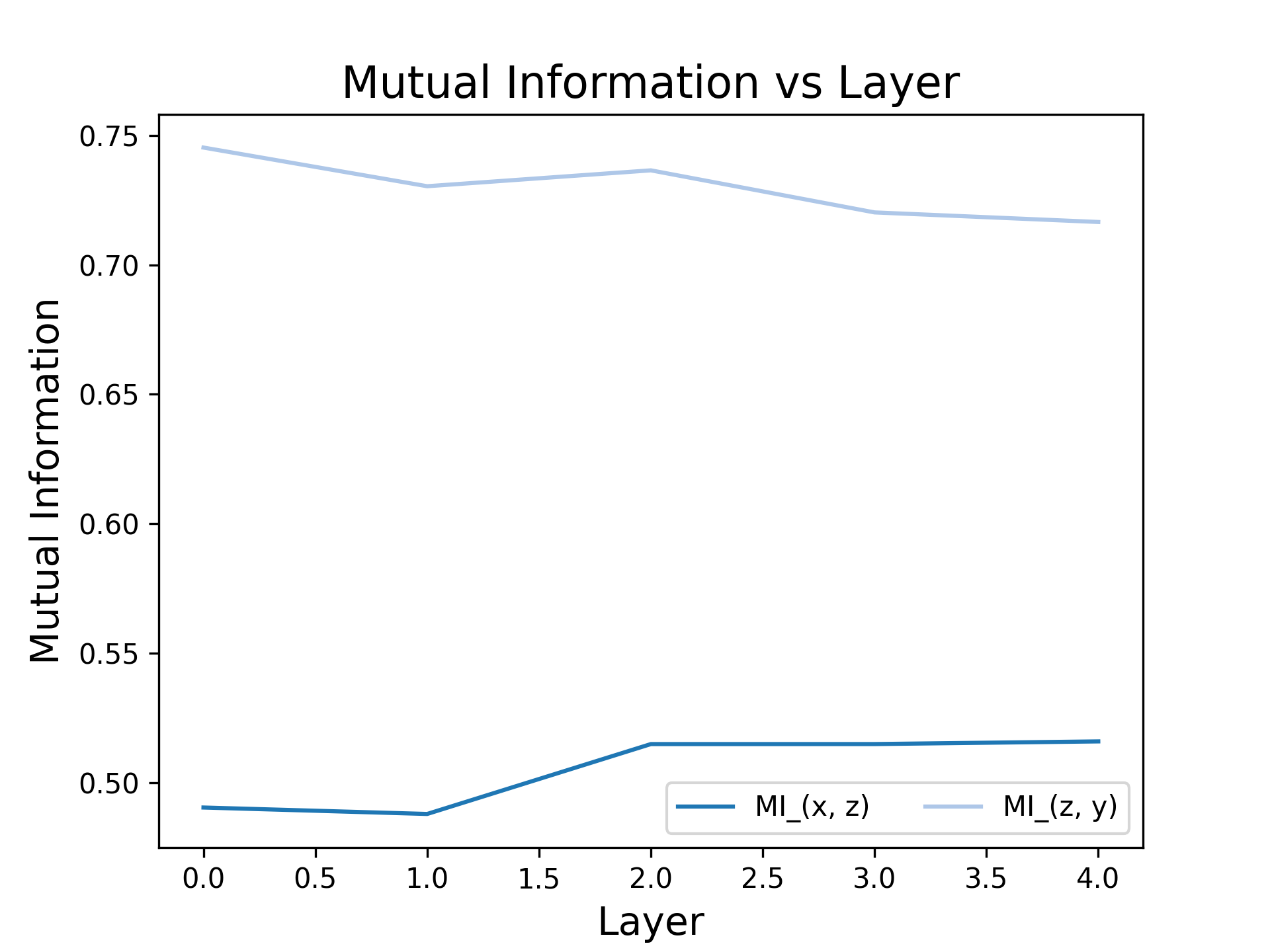}
\caption{\label{fig step_mi_mean}
Information Variation of LLMs in the Emotion Space with $d_k=32$
}
\end{figure}

\textbf{Objective}: Explore the decompression mechanism of LLMs.
Hypothesis: LLMs decompress information from the emotion space at the key time step 
$t$ to make predictions.

\textbf{Steps}:
We tested the mutual information between the LLMs' hidden state at time step 
$t$ and the hidden state at time step 0, as well as the mutual information with the emotion space $es^{top}_{d_k})$.
Here, the hidden state at time step 0 corresponds to the LLM sample 
$s_i$ 's hidden state representation in the emotion space, while the hidden states at other time steps correspond to the hidden representations during the generation process.
The objective of this experiment is to verify the information variation of LLMs at each time step during the generation process.

\begin{figure}
\centering
\includegraphics[width=70mm]{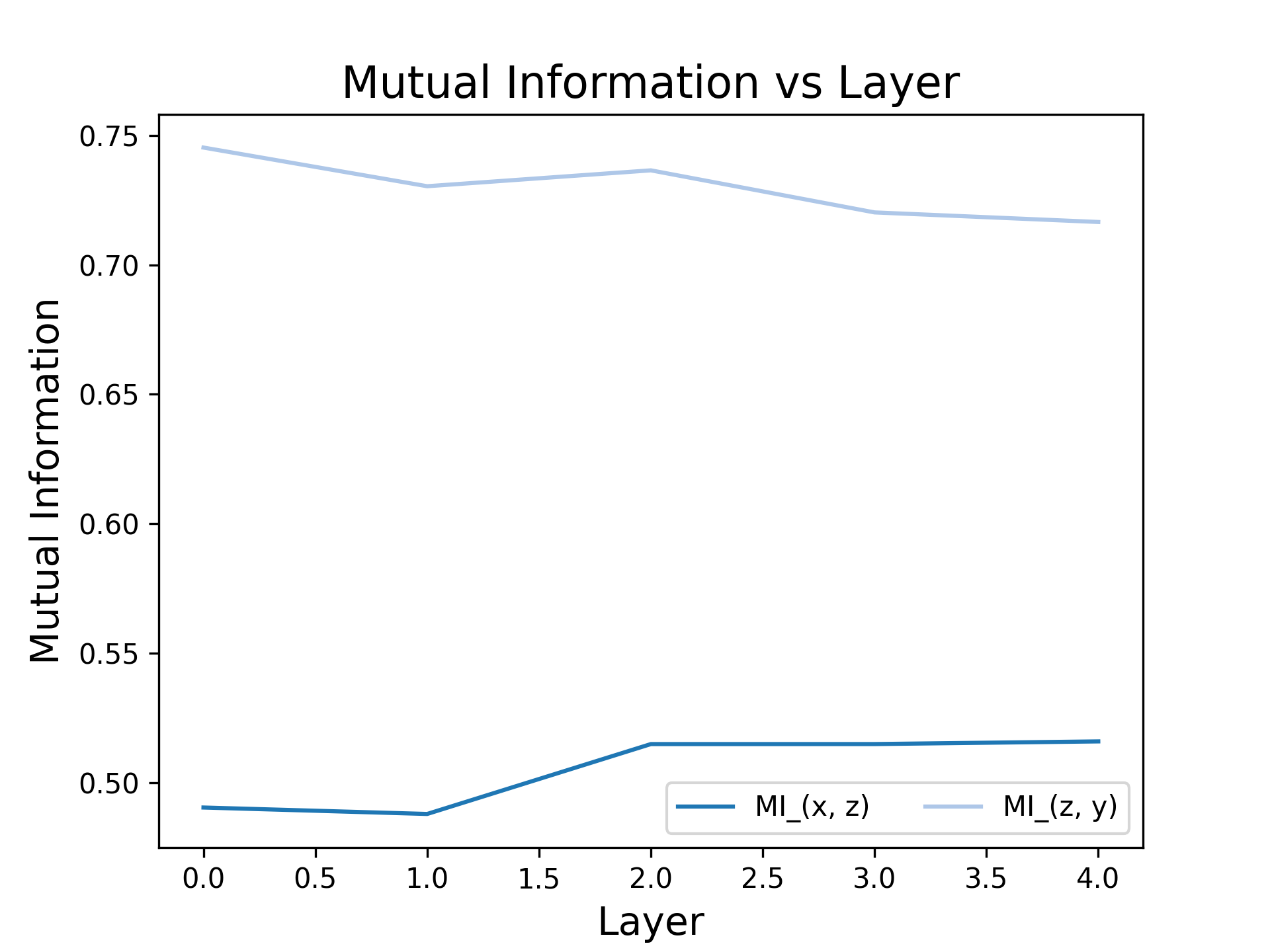}
\caption{\label{fig step_mi_total}
Information Variation of LLMs in the Emotion Space with $d_k=32$
}
\end{figure}

Figure \ref{fig step_mi_mean} shows the variation in mutual information between the LLM's hidden state at time step $t$ and the average emotion space of $s_i$ 's comprehension content.
Figure \ref{fig step_mi_total} shows the variation in mutual information between the LLM's hidden state at time step 
$t$ and the overall emotion space of 
$s_i$ 's comprehension content.
The results indicate that during the generation process, LLMs continuously decompress information from the emotion space to generate tokens.

\subsubsection{Detection of Key Prediction Steps}
To further investigate whether LLMs rely on certain key time steps 
$t$ during prediction, we conducted additional experiments.

\textbf{Objective}: Explore the key time steps during LLMs' prediction process.

\textbf{Hypothesis}: There exist key time steps $t$ that significantly affect the prediction results.

\textbf{Steps}:
At each time step 
$t$, we directly add or subtract the ground-truth emotion space to the LLM's hidden state to examine the effect of each time step.
The 0-th step represents the information representation during the understanding phase, while steps 1-5 represent the information representations during the generation process.

The experimental results are shown in Table 1. The results indicate that the most crucial time steps are the 0-th, 4-th, and 5-th steps. Modifying the LLM's hidden state at these time steps significantly altered the prediction accuracy. The impact of the 0-th step is mainly due to the fact that LLMs are processing and understanding the input sample 
$s_i$ at this point. The significant influence of the $4$-th and $5$-th steps is mainly because LLMs decompress key information at these moments.
\begin{table}
\centering
\begin{tabular}{ccccc}
\hline
\textbf{Steps} & \textbf{Add/Sub} & \textbf{Acc} & \textbf{F1}\\
\hline
% \multirow{3}{*}{\centering \shortstack{IAMM \\ vs. EmpDG}} & Emp. & \textbf{45.4} & 24.0 & 0.48\\
% & Rel. & \textbf{52.8} & 16.3 & 0.43 \\
% & Flu. & \textbf{50.1} & 5.9  & 0.45 \\

\hline

Baseline & - & 35.35 & 26.24\\
 \hline
\multirow{2}{*}{\centering \shortstack{Step 0}} & Addition &  & \\
 & Subtract  & 14.13 & 11.13\\
 \hline
\multirow{2}{*}{\centering \shortstack{Step 0-3}} & Addition &  & \\
 & Subtract  & 14.13 & 11.13\\
 \hline
 \multirow{2}{*}{\centering \shortstack{Step 4-5}} & Addition & 81.92 & 69.91\\
 & Subtract  & 11.41 & 7.81\\
 \hline
  \multirow{2}{*}{\centering \shortstack{Step 0,4-5}} & Addition  &  & \\
 & Subtract & 3.006 & 2.06\\
 \hline
\end{tabular}
\caption{\label{table human}
Experimental results of modifying hidden states at $t$ Time Steps.
}
\end{table}

In general, when LLMs are understanding the input information at $t$=0, removing the hidden state’s representation in the emotion space significantly reduces emotional accuracy. This is mainly because LLMs rely on decompressing the state at the understanding stage to make accurate predictions. Removing the hidden state’s representation in the emotion space at this point makes it difficult for the model to decompress effective information. On the other hand, adding the emotion space projection to the hidden state does not have a significant impact on the results, suggesting that LLMs have already effectively compressed information in the emotion space.

At the prediction stage ($t$=4,5), both adding and subtracting the hidden state’s representation in the emotion space significantly affect the results. This indicates that there are key moments during the prediction process when LLMs rely on specific emotional representations.

Furthermore, when the hidden state is added to the emotion space, the LLMs' performance greatly improves. This suggests that LLMs struggle to decompress high-quality information on their own, and better decompression requires additional strategies or support to effectively extract the desired information.

\section{Method}
To further enhance LLMs' ability to decompress information, we propose two models:
Information Compression-based Context Learning and Task-Space-guided Fine-Tuning.

\subsection{Information Compression-based Context Learning}

We first used a pre-trained model, $RoBERTa_{large}$, to retrieve the top $k_s$ most similar samples.
\begin{gather}
p_{s_i}, v_{s_i} = RoBERTa_{large}(s_i),\\
o_{m_i} = Cosine(v_{s_{i}}, v^s_{m_i}), m_i \in n_d,\\
s_j = Top_{k_2}(o_1, o_2, ..., o_{m_i}), j \in [1, k_2],
\label{equation 2}
\end{gather}

Next, we extracted the emotion labels of these $k_s$ samples and converted them into emotion vector representations. These vectors were then weighted by 
$w_e$ and added to the hidden layers of the LLMs.
\begin{gather}
    h^l_{e} = \sum^{k_1}_{n_i=1} w_e * h^l_{e_{n_i}} \\
    \widetilde{h}^l = h^l + h^l_{e}
\end{gather}

Intuitively, since the weight $w_e$ is fixed, the introduction of $k_2$  direction vectors, each added with equal weight, may not accurately provide the model with a high-quality hidden state. Therefore, we further refined the adjustment of these direction vectors.
\begin{gather}
    O^l_{E_i} = \widetilde{h}^l + h^l_{E_i} \\
    g_{E_i} = Softmax(\widetilde{h}^l  O^l_{E_i}) \\
    \overline{h}^l = \widetilde{h}^l + w^a_e  (g_{E_i} h^l_{E_i})
\end{gather}

Finally, we prompt the LLM to predict the emotion category of the dialogue context.
\begin{gather}
    e_i = LLM(s_i) \\
    Y = LLM(y_t|s_i, e_i)
\end{gather}

\subsection{Task-Space-guided Fine-Tuning}

For the sample $s_i$, this section transforms its corresponding emotion label $e^*$ into the direction vector at the $l$-th layer. Then, it is added to the hidden layer of the LLMs at the $l$-th layer to obtain a higher-quality hidden layer.
\begin{gather}
    h^{*,l} = h^l + h^l_{e^*}
\end{gather}

To ensure that the hidden layers converge towards a better-quality direction during training, we design a mean squared error (MSE) loss.
\begin{gather}
    \mathcal{L}^l_{mse} = \frac{1}{d} \sum^d_{i=1} (h^{*,l}_i - h^l)^2 \\
    \mathcal{L}_{mse} = w_{mse} * \sum^L_{l=1} \mathcal{L}^l_{mse}
\end{gather}
where $w_{mse}$ is a hyperparameter, and $n$ and $L$ are the dimensionality and number of layers of the LLMs, respectively.

For $s_i$, we also employ cross-entropy as the generation loss to encourage the LLMs to produce outputs in the corresponding response format.
%: ``Emotion: [predicted emotion type]''
\begin{equation}
    \mathcal{L}_{\text{LM}} = -\sum_{t=1}^{T} \log p_\theta(x_t | x_{<t}, s_i)
\end{equation}
where $x_{<t}$ is the previously generated text. Curriculum learning is used for optimization in this case. $T$ represents the training time step.

Overall, we optimize the model based on the two losses:
\begin{gather}
    \mathcal{L} = \mathcal{L}_{mse} + \mathcal{L}_{\text{LM}}
\end{gather}

% \section{Limitations}
\section{Results and Analysis}

We validate the proposed method on the Empathetic Dialogues dataset, with the results shown in Table \ref{table result}.
The results indicate that the performance of the proposed IC-ICL method significantly outperforms the baseline. Along with improving inference performance, the inference speed also shows a substantial improvement.
At the same time, the proposed TS-FT method also outperforms the baseline. The advantages of both methods show that enhancing the information decompression capability of LLMs further promotes their performance.

\section{Related Work}
Large language models have proven effective in various tasks, such as sentiment analysis, semantic parsing~\cite{song2019hierarchical}, and logical reasoning. Current research primarily focuses on enhancing these models through prompting or training, with in-context learning and fine-tuning being the most representative and widely used methods.

\textbf{In-Context Learning}. Wei et al.~\cite{wei2022chain} propose chain-of-thought prompting, which decomposes reasoning into sequential logical steps to enhance LLMs' structured reasoning.
Wang et al.~\cite{Wang2022SelfConsistencyIC} introduce self-consistency prompting, generating multiple reasoning paths and selecting the final answer through majority voting.
Yao et al.~\cite{Yao2023TreeOT} develop a tree-based approach that breaks complex problems into hierarchical sub-problems to improve reasoning accuracy.
Besta et al.~\cite{Besta2023GraphOT} propose a graph-based reasoning method that uses feedback loops to iteratively refine reasoning quality.

These methods rely on manually constructed examples, limiting their generalizability to diverse tasks.
To address this, retrieval-based approaches select examples based on lexical features~\cite{rubin2021learning,agrawal2022context,luo2023dr}, semantic similarity~\cite{Liu2021WhatMG}, structural patterns~\cite{levy2022diverse}, or other factors~\cite{fu2022complexity,gonen2022demystifying,drozdov2022compositional}.
While these approaches show promising performance, they significantly slow down the inference process of LLMs due to the increased input length caused by the additional examples.

\textbf{Fine-tuning}.
Fine-tuning is a key approach for adapting large language models (LLMs) to downstream tasks. While full model fine-tuning updates all parameters to achieve task-specific objectives~\cite{devlin2019bert, radford2019language}, it is computationally expensive.
To improve efficiency, parameter-efficient methods have been proposed. Adapter-tuning adds small task-specific layers while freezing the main model~\cite{houlsby2019parameter, pfeiffer2020adapterfusion}. Prefix-tuning optimizes prefix vectors prepended to each layer's input~\cite{li2021prefix, lester2021power}, and LoRA modifies low-rank matrices within the model~\cite{hu2021lora}.

Other strategies include prompt-tuning, which fine-tunes continuous prompt embeddings~\cite{liu2021ptuning}, and task-specific loss functions, such as contrastive or reinforcement learning objectives~\cite{gao2021simcse, ouyang2022training}. These methods improve performance and efficiency, enabling the practical deployment of LLMs.

\begin{table}
\centering
\begin{tabular}{ccccc}
\hline
\textbf{Type} & \textbf{Models} & \textbf{Acc} & \textbf{F1} & \textbf{T}\\
 \hline
\multirow{3}{*}{\centering \shortstack{Prompt}} & Llama3.1$_{8b}$ & 35.35 & 26.24 & 7:12\\
 & ICL & 29.55 & 32.45 & 14:17\\
 & IC-ICL & \textbf{43.69} & \textbf{36.35} & 7:37\\
 \hline
 \multirow{2}{*}{\centering \shortstack{Fine-Tuning}} & Llama3.1$_{8b}$ & 54.48 & 53.94 & - \\
 & TS-FT & \textbf{55.43} & \textbf{54.96} & -\\
 \hline
\end{tabular}
\caption{\label{table result}
Experimental results.
}
\end{table}

\section{Conclusion}
This paper investigates the information processing mechanisms of Large Language Models (LLMs) through the lens of Information Bottleneck Theory. We show that LLMs compress input into task-specific spaces but struggle with decompression during prediction. Based on these insights, we propose two methods: Information Compression-based Context Learning (IC-ICL) and Task-Space-guided Fine-Tuning (TS-FT).

IC-ICL improves reasoning accuracy and accelerates inference by over 40\%, while TS-FT enhances decompression capabilities through a simple loss function. Our experiments validate the effectiveness of these approaches, demonstrating significant performance improvements. These findings offer a deeper understanding of LLMs' information processing and provide practical solutions for enhancing model performance.

Future work will explore refining these methods and applying them to other NLP tasks, further enhancing the efficiency and accuracy of LLMs.

\section*{Ethical Considerations}
Regarding the potential ethical impacts of our work:
(1) The dataset we use is EMPATHETIC-DIALOGUE, which is open source and does not involve any potential ethical risks.
(2) The baseline models we use are also public and do not have potential moral impacts. 
Moreover, the components employed in our model are open-sourced or innovative and do not involve potential ethical risks.

\section*{Acknowledgments}
We are grateful to the reviewers for their diligent evaluation and constructive feedback, which helped enhance the quality of this paper. 
We also appreciate the insightful discussions and comments from the authors, which stimulated valuable thinking and contributed significantly to the development of this research.
This work was supported by National Natural Science Foundation of China (No.61976054).

%\clearpage

\bibliography{custom}

\begin{thebibliography}{24}
\expandafter\ifx\csname natexlab\endcsname\relax\def\natexlab#1{#1}\fi

\bibitem[{Agrawal et~al.(2022)Agrawal, Zhou, Lewis, Zettlemoyer, and Ghazvininejad}]{agrawal2022context}
Sweta Agrawal, Chunting Zhou, Mike Lewis, Luke Zettlemoyer, and Marjan Ghazvininejad. 2022.
\newblock In-context examples selection for machine translation.
\newblock \emph{arXiv preprint arXiv:2212.02437}.

\bibitem[{Besta et~al.(2023)Besta, Blach, Kub\'{\i}\v{c}ek, Gerstenberger, Gianinazzi, Gajda, Lehmann, Podstawski, Niewiadomski, Nyczyk, and Hoefler}]{Besta2023GraphOT}
Maciej Besta, Nils Blach, Ale\v{s} Kub\'{\i}\v{c}ek, Robert Gerstenberger, Lukas Gianinazzi, Joanna Gajda, Tomasz Lehmann, Michal Podstawski, Hubert Niewiadomski, Piotr Nyczyk, and Torsten Hoefler. 2023.
\newblock \href {https://api.semanticscholar.org/CorpusID:261030303} {Graph of thoughts: Solving elaborate problems with large language models}.
\newblock In \emph{AAAI Conference on Artificial Intelligence}.

\bibitem[{Devlin et~al.(2019)Devlin, Chang, Lee, and Toutanova}]{devlin2019bert}
Jacob Devlin, Ming-Wei Chang, Kenton Lee, and Kristina Toutanova. 2019.
\newblock \href {https://api.semanticscholar.org/CorpusID:52967399} {Bert: Pre-training of deep bidirectional transformers for language understanding}.
\newblock In \emph{North American Chapter of the Association for Computational Linguistics}.

\bibitem[{Drozdov et~al.(2022)Drozdov, Sch{\"a}rli, Aky{\"u}rek, Scales, Song, Chen, Bousquet, and Zhou}]{drozdov2022compositional}
Andrew Drozdov, Nathanael Sch{\"a}rli, Ekin Aky{\"u}rek, Nathan Scales, Xinying Song, Xinyun Chen, Olivier Bousquet, and Denny Zhou. 2022.
\newblock Compositional semantic parsing with large language models.
\newblock In \emph{The Eleventh International Conference on Learning Representations}.

\bibitem[{Fu et~al.(2022)Fu, Peng, Sabharwal, Clark, and Khot}]{fu2022complexity}
Yao Fu, Hao Peng, Ashish Sabharwal, Peter Clark, and Tushar Khot. 2022.
\newblock Complexity-based prompting for multi-step reasoning.
\newblock In \emph{The Eleventh International Conference on Learning Representations}.

\bibitem[{Gao et~al.(2021)Gao, Yao, and Chen}]{gao2021simcse}
Tianyu Gao, Xingcheng Yao, and Danqi Chen. 2021.
\newblock \href {https://api.semanticscholar.org/CorpusID:233296292} {Simcse: Simple contrastive learning of sentence embeddings}.
\newblock \emph{ArXiv}, abs/2104.08821.

\bibitem[{Gonen et~al.(2022)Gonen, Iyer, Blevins, Smith, and Zettlemoyer}]{gonen2022demystifying}
Hila Gonen, Srini Iyer, Terra Blevins, Noah~A Smith, and Luke Zettlemoyer. 2022.
\newblock Demystifying prompts in language models via perplexity estimation.
\newblock \emph{arXiv preprint arXiv:2212.04037}.

\bibitem[{Houlsby et~al.(2019)Houlsby, Giurgiu, Jastrzebski, Morrone, de~Laroussilhe, Gesmundo, Attariyan, and Gelly}]{houlsby2019parameter}
Neil Houlsby, Andrei Giurgiu, Stanislaw Jastrzebski, Bruna Morrone, Quentin de~Laroussilhe, Andrea Gesmundo, Mona Attariyan, and Sylvain Gelly. 2019.
\newblock \href {https://api.semanticscholar.org/CorpusID:59599816} {Parameter-efficient transfer learning for nlp}.
\newblock \emph{ArXiv}, abs/1902.00751.

\bibitem[{Hu et~al.(2021)Hu, Shen, Wallis, Allen-Zhu, Li, Wang, and Chen}]{hu2021lora}
J.~Edward Hu, Yelong Shen, Phillip Wallis, Zeyuan Allen-Zhu, Yuanzhi Li, Shean Wang, and Weizhu Chen. 2021.
\newblock \href {https://api.semanticscholar.org/CorpusID:235458009} {Lora: Low-rank adaptation of large language models}.
\newblock \emph{ArXiv}, abs/2106.09685.

\bibitem[{Lester et~al.(2021)Lester, Al-Rfou, and Constant}]{lester2021power}
Brian Lester, Rami Al-Rfou, and Noah Constant. 2021.
\newblock \href {https://api.semanticscholar.org/CorpusID:233296808} {The power of scale for parameter-efficient prompt tuning}.
\newblock In \emph{Conference on Empirical Methods in Natural Language Processing}.

\bibitem[{Levy et~al.(2022)Levy, Bogin, and Berant}]{levy2022diverse}
Itay Levy, Ben Bogin, and Jonathan Berant. 2022.
\newblock Diverse demonstrations improve in-context compositional generalization.
\newblock \emph{arXiv preprint arXiv:2212.06800}.

\bibitem[{Li and Liang(2021)}]{li2021prefix}
Xiang~Lisa Li and Percy Liang. 2021.
\newblock \href {https://api.semanticscholar.org/CorpusID:230433941} {Prefix-tuning: Optimizing continuous prompts for generation}.
\newblock \emph{Proceedings of the 59th Annual Meeting of the Association for Computational Linguistics and the 11th International Joint Conference on Natural Language Processing (Volume 1: Long Papers)}, pages 4582--4597.

\bibitem[{Liu et~al.(2024)Liu, Zhang, Guo, Wang, Dong, Li, Lee, Zhang, and Liu}]{Liu2024CtrlAAR}
Huanshuo Liu, Hao Zhang, Zhijiang Guo, Jing Wang, Kuicai Dong, Xiangyang Li, Yi~Lee, Cong Zhang, and Yong Liu. 2024.
\newblock \href {https://api.semanticscholar.org/CorpusID:273163564} {Ctrla: Adaptive retrieval-augmented generation via inherent control}.
\newblock \emph{arXiv}.

\bibitem[{Liu et~al.(2021{\natexlab{a}})Liu, Shen, Zhang, Dolan, Carin, and Chen}]{Liu2021WhatMG}
Jiachang Liu, Dinghan Shen, Yizhe Zhang, Bill Dolan, Lawrence Carin, and Weizhu Chen. 2021{\natexlab{a}}.
\newblock \href {https://api.semanticscholar.org/CorpusID:231632658} {What makes good in-context examples for gpt-3?}
\newblock In \emph{Workshop on Knowledge Extraction and Integration for Deep Learning Architectures; Deep Learning Inside Out}.

\bibitem[{Liu et~al.(2021{\natexlab{b}})Liu, Ji, Fu, Du, Yang, and Tang}]{liu2021ptuning}
Xiao Liu, Kaixuan Ji, Yicheng Fu, Zhengxiao Du, Zhilin Yang, and Jie Tang. 2021{\natexlab{b}}.
\newblock \href {https://api.semanticscholar.org/CorpusID:238857040} {P-tuning v2: Prompt tuning can be comparable to fine-tuning universally across scales and tasks}.
\newblock \emph{ArXiv}, abs/2110.07602.

\bibitem[{Luo et~al.(2023)Luo, Xu, Dai, Pasupat, Kazemi, Baral, Imbrasaite, and Zhao}]{luo2023dr}
Man Luo, Xin Xu, Zhuyun Dai, Panupong Pasupat, Mehran Kazemi, Chitta Baral, Vaiva Imbrasaite, and Vincent~Y Zhao. 2023.
\newblock Dr. icl: Demonstration-retrieved in-context learning.
\newblock \emph{arXiv preprint arXiv:2305.14128}.

\bibitem[{Ouyang et~al.(2022)Ouyang, Wu, Jiang, Almeida, Wainwright, Mishkin, Zhang, Agarwal, Slama, Ray, Schulman, Hilton, Kelton, Miller, Simens, Askell, Welinder, Christiano, Leike, and Lowe}]{ouyang2022training}
Long Ouyang, Jeff Wu, Xu~Jiang, Diogo Almeida, Carroll~L. Wainwright, Pamela Mishkin, Chong Zhang, Sandhini Agarwal, Katarina Slama, Alex Ray, John Schulman, Jacob Hilton, Fraser Kelton, Luke~E. Miller, Maddie Simens, Amanda Askell, Peter Welinder, Paul~Francis Christiano, Jan Leike, and Ryan~J. Lowe. 2022.
\newblock \href {https://api.semanticscholar.org/CorpusID:246426909} {Training language models to follow instructions with human feedback}.
\newblock \emph{ArXiv}, abs/2203.02155.

\bibitem[{Pfeiffer et~al.(2020)Pfeiffer, Kamath, R{\"u}ckl{\'e}, Cho, and Gurevych}]{pfeiffer2020adapterfusion}
Jonas Pfeiffer, Aishwarya Kamath, Andreas R{\"u}ckl{\'e}, Kyunghyun Cho, and Iryna Gurevych. 2020.
\newblock \href {https://api.semanticscholar.org/CorpusID:218470208} {Adapterfusion: Non-destructive task composition for transfer learning}.
\newblock \emph{ArXiv}, abs/2005.00247.

\bibitem[{Radford et~al.(2019)Radford, Wu, Child, Luan, Amodei, Sutskever et~al.}]{radford2019language}
Alec Radford, Jeffrey Wu, Rewon Child, David Luan, Dario Amodei, Ilya Sutskever, et~al. 2019.
\newblock Language models are unsupervised multitask learners.
\newblock \emph{OpenAI blog}, 1(8):9.

\bibitem[{Rubin et~al.(2021)Rubin, Herzig, and Berant}]{rubin2021learning}
Ohad Rubin, Jonathan Herzig, and Jonathan Berant. 2021.
\newblock Learning to retrieve prompts for in-context learning.
\newblock \emph{arXiv preprint arXiv:2112.08633}.

\bibitem[{Song et~al.(2019)Song, Zhan, and Haihong}]{song2019hierarchical}
Meina Song, Zecheng Zhan, and E~Haihong. 2019.
\newblock Hierarchical schema representation for text-to-sql parsing with decomposing decoding.
\newblock \emph{IEEE Access}, 7:103706--103715.

\bibitem[{Wang et~al.(2022)Wang, Wei, Schuurmans, Le, Chi, and Zhou}]{Wang2022SelfConsistencyIC}
Xuezhi Wang, Jason Wei, Dale Schuurmans, Quoc Le, Ed~H. Chi, and Denny Zhou. 2022.
\newblock \href {https://api.semanticscholar.org/CorpusID:247595263} {Self-consistency improves chain of thought reasoning in language models}.
\newblock \emph{ArXiv}, abs/2203.11171.

\bibitem[{Wei et~al.(2022)Wei, Wang, Schuurmans, Bosma, Xia, Chi, Le, Zhou et~al.}]{wei2022chain}
Jason Wei, Xuezhi Wang, Dale Schuurmans, Maarten Bosma, Fei Xia, Ed~Chi, Quoc~V Le, Denny Zhou, et~al. 2022.
\newblock Chain-of-thought prompting elicits reasoning in large language models.
\newblock \emph{Advances in neural information processing systems}, 35:24824--24837.

\bibitem[{Yao et~al.(2023)Yao, Yu, Zhao, Shafran, Griffiths, Cao, and Narasimhan}]{Yao2023TreeOT}
Shunyu Yao, Dian Yu, Jeffrey Zhao, Izhak Shafran, Thomas~L. Griffiths, Yuan Cao, and Karthik Narasimhan. 2023.
\newblock \href {https://api.semanticscholar.org/CorpusID:258762525} {Tree of thoughts: Deliberate problem solving with large language models}.
\newblock \emph{ArXiv}, abs/2305.10601.

\end{thebibliography}
\bibliographystyle{acl_natbib}

\appendix

\end{document}